\documentclass{article}

\usepackage[preprint, nonatbib]{paper_style}

\usepackage{tabularx}
\usepackage{graphicx}
\usepackage{amsmath}
\usepackage[utf8]{inputenc} 
\usepackage[T1]{fontenc}    
\usepackage[hidelinks]{hyperref}       
\usepackage{url}            
\usepackage{booktabs}       
\usepackage{amsfonts}       
\usepackage{nicefrac}       
\usepackage{microtype}      
\usepackage{xcolor}         
\title{Kolmogorov-Arnold Network Autoencoders}

\author{%
  Mohammadamin~Moradi \\
  Arizona State University\\
  Tempe, AZ 85201 \\
  \texttt{mmoradi5@asu.edu} \\
  \And
  Shirin Panahi \\
  Arizona State University\\
  Tempe, AZ 85201 \\
  \texttt{spanahi1@asu.edu} \\
  \And
  Erik Bollt \\
  Clarkson University\\
  Potsdam, NY 13699 \\
  \texttt{ebollt@clarkson.edu} \\
  \And
  Ying-Cheng Lai \\
  Arizona State University\\
  Tempe, AZ 85201 \\
  \texttt{Ying-Cheng.Lai@asu.edu} \\
}

\begin{document}
\maketitle

\begin{abstract}

Deep learning models have revolutionized various domains, with Multi-Layer Perceptrons (MLPs) being a cornerstone for tasks like data regression and image classification. However, a recent study has introduced Kolmogorov-Arnold Networks (KANs) as promising alternatives to MLPs, leveraging activation functions placed on edges rather than nodes. This structural shift aligns KANs closely with the Kolmogorov-Arnold representation theorem, potentially enhancing both model accuracy and interpretability. In this study, we explore the efficacy of KANs in the context of data representation via autoencoders, comparing their performance with traditional Convolutional Neural Networks (CNNs) on the MNIST, SVHN, and CIFAR-10 datasets. Our results demonstrate that KAN-based autoencoders achieve competitive performance in terms of reconstruction accuracy, thereby suggesting their viability as effective tools in data analysis tasks. 
\end{abstract}

\section{Introduction}

In recent years, deep learning has seen unprecedented advancements, catalyzing breakthroughs in various fields such as image recognition, natural language processing, medical diagnostics, and cybersecurity~\cite{moradi2022defending, moradi2024heterogeneous, moradi2024random, zhai2023model, wu2015image, goldberg2017neural, ker2017deep, li2022research}. Central to these advancements are neural network architectures like MLPs, which have proven effective in tasks ranging from simple regression to complex image classification. MLPs are well-known for their ability to approximate a wide range of functions, a property substantiated by the universal approximation theorem~\cite{baker1998universal, lu2020universal, voigtlaender2023universal}. However, deep learning architectures are constantly evolving, driven by the constant thirst for improved performance, interpretability, and efficiency. One of the emerging paradigms challenging the dominance of MLPs is the recently introduced KANs~\cite{liu2024kan, liu2024kan2.0}. KANs draw inspiration from the Kolmogorov-Arnold representation theorem, which asserts that any continuous multivariate function can be represented as a composition of functions of a single variable. This theorem underpins the structure of KANs, where activation functions are trained and applied not at neuron nodes but directly on the edges of the network graph. This alters the dynamics of information flow within the network and presents intriguing implications: KANs potentially offer enhanced model capacity, better handling of complex dependencies in data, and improved interpretability of learned representations compared to traditional MLPs. By decentralizing activation functions to the edges, KANs facilitate a more modular approach to feature extraction and transformation, potentially yielding more structured and interpretable representations of input data.

KAN has received a great deal of attention recently. MonoKAN builds on the KAN architecture and is a version that enforces certified partial monotonicity. This modification ensures that model predictions align with expert-imposed monotonicity constraints, making it highly interpretable and suitable for applications requiring transparent, explainable AI. MonoKAN leverages cubic Hermite splines to guarantee monotonicity while improving predictive performance over existing monotonic MLP approaches~\cite{polomolina2024monokancertifiedmonotonickolmogorovarnold}. Another paper compared KANs to MLPs in low-data environments. KANs' reliance on complex, learnable activation functions increases parameter count, which becomes a bottleneck when training on small datasets. In such cases, MLPs with custom activation functions outperformed KANs significantly, especially when only a few hundred samples were available~\cite{pourkamalianaraki2024kolmogorovarnoldnetworkslowdataregimes}. Moreover, to address KANs' computational inefficiency, researchers proposed FastKAN, a version that replaces B-splines with Gaussian radial basis functions. This approach drastically improves training speed (by over three times) without sacrificing accuracy, particularly in benchmark tasks like MNIST classification. This shows that KANs can be simplified while maintaining performance~\cite{li2024kolmogorovarnoldnetworksradialbasis}. One significant development is the integration of KANs into transfer learning. A recent paper suggests replacing the traditional linear probing layer in transfer learning with a KAN-based approach to model more complex relationships between data points. By using KANs in combination with a pre-trained ResNet-50 model, the researchers demonstrated improvements in accuracy and generalization, particularly on the CIFAR-10 dataset~\cite{shen2024reimagininglinearprobingkolmogorovarnold}. Another notable contribution is the introduction of the Kolmogorov-Arnold Transformer (KAT). This model replaces the MLP layers in standard transformers with KAN layers, making it more expressive. The researchers behind KAT addressed key challenges like computational inefficiency and weight initialization by implementing rational functions and variance-preserving initialization techniques. This architecture shows promising results, especially in scaling KAN-based transformers for modern GPU hardware~\cite{yang2024kolmogorovarnoldtransformer}. Moroever, in Ref.~\cite{mehrabian2024implicitneuralrepresentationsfourier} authors focused on enhancing Implicit Neural Representations (INRs) by integrating Fourier series-based activation functions to better capture task-specific frequency components. The study shows that Fourier KANs (FKAN) improve on baseline models in terms of various metrics in both image representation and 3D occupancy tasks.

Another key development is the introduction of convolutional layers into KANs, which enhances their capacity to handle spatial information, as seen in experiments on the MNIST and Fashion MNIST datasets. These models demonstrated higher accuracy compared to small CNNs and performed slightly below medium CNNs, showing KANs' competitive advantage in image recognition tasks while keeping parameter complexity low by reducing the need for fully connected layers~\cite{bodner2024convolutional}. Motivated by these advancements, this study explores the application of KANs in the context of autoencoders for image representation tasks. Autoencoders are neural network architectures designed for unsupervised learning, tasked with learning efficient representations of input data through an encoder-decoder framework~\cite{bank2023autoencoders, pinaya2020autoencoders}. Traditionally, Convolutional Neural Networks (CNNs) have dominated in tasks involving image data, owing to their ability to capture spatial hierarchies and translational invariance~\cite{jana2022cnn}. In contrast to CNNs, KAN-based autoencoders leverage edge-based activations to potentially capture more nuanced relationships and dependencies within images. This suggests a novel perspective on how neural networks can be structured and optimized for image representation tasks. There are multiple projects and studies in progress regarding KAN autoencoders. For instance, NeuroBender combines the KAN with a Wasserstein Auto-Encoder to provide a unique approach to image processing on the MNIST dataset~\cite{hryszko2023kanneurobender}. Another project investigates the potential of KAN to represent the sinusoidal and other complicated signals~\cite{rong2023kanautoencoder}. Moreover, Ref.~\cite{wang2024kanbasedautoencodersfactor} uses KAN approach to latent factor conditional asset pricing models. This study aims to investigate whether KAN autoencoders can achieve comparable or superior performance on benchmark datasets.

\section{Methods} \label{sec:methods}

\subsection{Autoencoders and Features Reduction} \label{sec:meth_ae}

The primary objective of an autoencoder is to map input data into a lower-dimensional latent space (encoding) and then reconstruct the original input from this compressed representation (decoding). This process involves two main components: the encoder and the decoder. The encoding is achieved through a series of neural network layers that reduce the dimensionality of the input while retaining essential features. Mathematically, if $\mathbf{x}$ represents the input, the encoder function $f_{\theta}$ transforms $\mathbf{x}$ into a latent representation $\mathbf{z}$ (a.k.a. the ``bottleneck''), where $\mathbf{z} = f_{\theta}(\mathbf{x})$. On the other hand, the decoder reconstructs the input data from the latent representation. The decoder function \( g_{\phi} \) aims to reverse the encoding process, producing a reconstruction \( \mathbf{\hat{x}} \) from \( \mathbf{z} \), such that \( \mathbf{\hat{x}} = g_{\phi}(\mathbf{z}) \).

The loss function commonly used for training autoencoders is the Mean Squared Error (MSE) between the input \( \mathbf{x} \) and the reconstructed output \( \mathbf{\hat{x}} \):
\begin{align} \label{eq:mse}
\mathcal{L}(\mathbf{x}, \mathbf{\hat{x}}) = \| \mathbf{x} - \mathbf{\hat{x}} \|^2    
\end{align}

Traditional autoencoders leverage fully connected layers or convolutional layers, particularly when dealing with image data. Convolutional Autoencoders are a specialized form designed to exploit the spatial hierarchies in image data where the encoder consists of convolutional layers that downsample the input, followed by a bottleneck layer that represents the latent space. The decoder, in turn, uses convolutional transpose layers to upsample the latent representation and reconstruct the original image~\cite{alqahtani2018deep}.

\subsection{Kolmogorov-Arnold Networks (KAN)} \label{sec:meth_kan}

KANs represent an innovative approach to neural network architecture, inspired by the Kolmogorov-Arnold representation theorem. This theorem asserts that any continuous multivariate function can be expressed as a composition of continuous functions of a single variable and addition operations. KANs leverage this theorem by placing activation functions on the edges of the network graph rather than at the nodes, as is typical in MLPs and CNNs. This means that each connection between neurons includes an activation function, which transforms the information flow between nodes. Also, by decentralizing the activation functions to the edges, KANs promote a more modular approach to feature extraction and transformation. This modularity potentially enhances the interpretability and flexibility of the network, allowing for more nuanced representations of the input data.

As mentioned above, the Kolmogorov-Arnold representation theorem states that any continuous multivariate function $f: \mathbb{R}^n \to \mathbb{R}$ can be represented as a superposition of continuous univariate functions. The general form of the theorem is:

\begin{align}
f(x_1, x_2, \dots, x_n) = \sum_{i=1}^{2n-1} \phi_i \left( \sum_{j=1}^{n} \psi_{ij}(x_j) \right)
\end{align}

where $\phi_i$ and $\psi_{ij}$ are continuous univariate functions. Each of these functions is parameterized using a linear combination of the residual and spline functions. Spline functions are piecewise polynomials that can adapt to the data more flexibly than traditional linear models and are parametrized as a linear combination of B-splines such that:

\begin{align}
S(x) = \sum_{i=0}^{n} c_i B_{i,d}(x).
\end{align}

where $c_i$ are the trainable coefficients and $B_{i,d}(x)$ are the B-spline basis functions of degree $d$~\cite{bsplines}. KANs scale more efficiently than traditional MLPs, which typically scale as $O(n^2)$ with the number of neurons and layers. The spline-based structure of KANs allows them to achieve high accuracy with fewer parameters. The empirical results suggest that KANs demonstrate faster scaling behavior compared to MLPs, leading to improved accuracy and efficiency in tasks like data fitting and solving partial differential equations (PDEs)~\cite{liu2024kan, liu2024kan2.0}.

\subsection{KAN Autoencoders} \label{sec:meth_kan}

The encoder-decoder structure of KAN-based autoencoders mirrors that of traditional autoencoders but with the key difference of edge-based activations. This architectural shift aims to capture more complex dependencies within the data, potentially leading to better performance and more interpretable latent representations. In this study, we integrate KANs into the autoencoder framework to evaluate their efficacy in image representation tasks. We compare the performance of KAN-based autoencoders with that of traditional convolutional autoencoders on the MNIST, SVHN, and CIFAR-10 datasets, assessing both reconstruction accuracy and the quality of the learned features.

\begin{figure}
    \centering
    \includegraphics[width=1\linewidth]{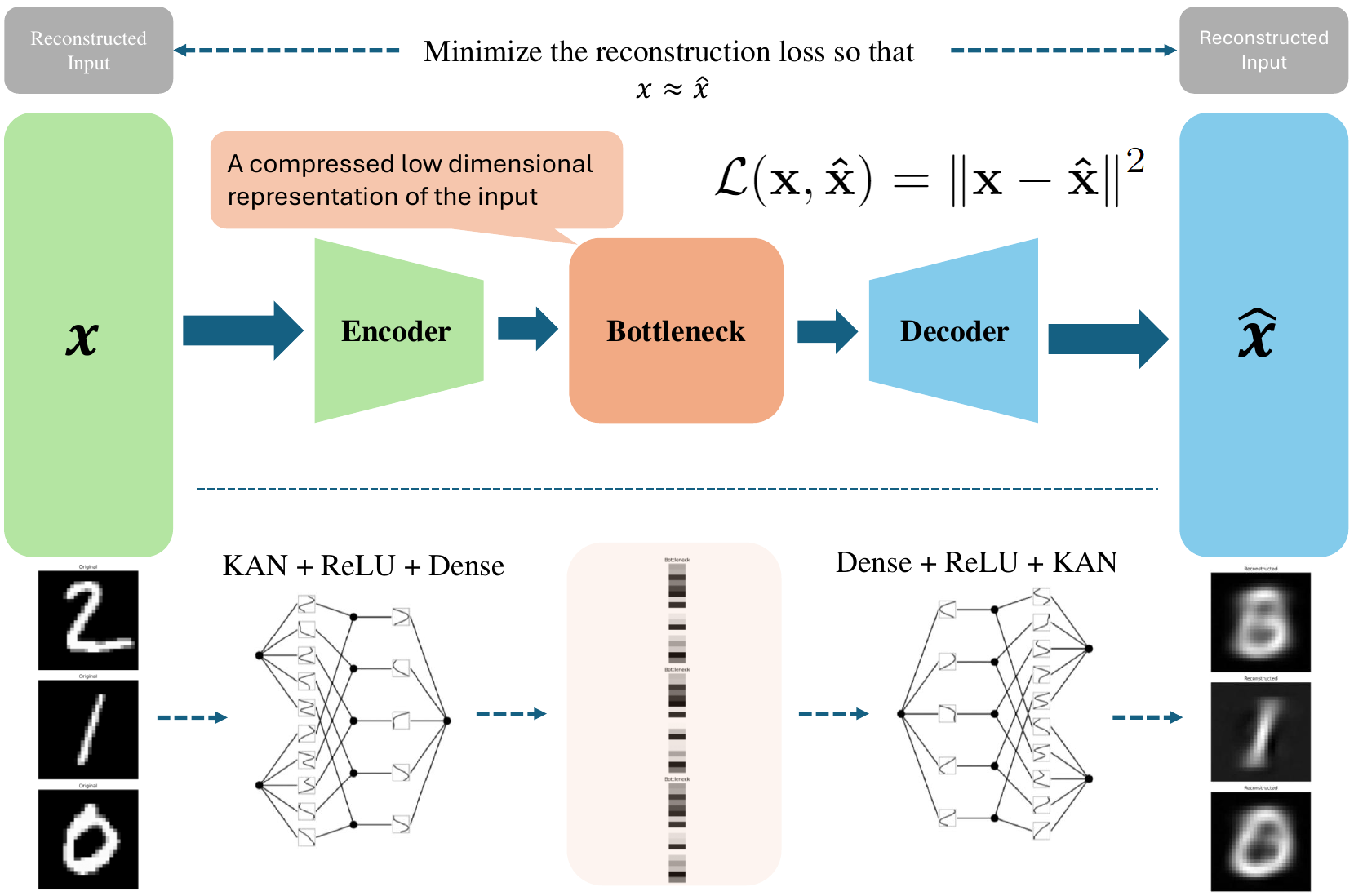}
    \caption{KAN Autoencoder Structure. The structure of our KAN autoencoder consists of an encoder and a decoder. The encoder includes a KAN layer, a ReLU activation, and a dense layer, which transforms the input size to a hidden size and then to the bottleneck size. For example, it maps from 784 to 8, followed by a ReLU activation, and then from 8 to 18. The decoder reverses this process, starting with a dense layer, followed by a ReLU activation, and finally a KAN layer, mapping from the bottleneck size back to the hidden size and the original input size, i.e., from 18 to 8, followed by ReLU, and from 8 to 784.}
    \label{fig:structure}
\end{figure}

Let the input image be $x \in \mathbb{R}^{h \times w}$, where $h$ and $w$ are the height and width of the image, respectively. The encoder is represented as:

\begin{align}
z = f_{\text{Encoder}}(x) = W_{\text{dense}} \cdot \max(0, f_{\text{KAN\_1}}(x)) + b_{\text{dense}}.
\end{align}

The latent representation $z \in \mathbb{R}^d$ captures the essential features of the input in a compressed form, where $d \ll h \times w$. The latent vector $z$ is then passed to the decoder for reconstruction. The decoder reconstructs the input image $\hat{x}$ from the latent vector $z$ afterwards.
The decoder can be described as:

\begin{align}
\hat{x}  = f_{\text{Decoder}}(z) = f_{\text{KAN\_2}}(\max(0, W'_{\text{dense}} \cdot z + b'_{\text{dense}})).
\end{align}

The autoencoder is trained by minimizing the reconstruction error between the original image $x$ and the reconstructed image $\hat{x}$. The loss function is the MSE (see Eq.~\ref{eq:mse}). The use of KAN layers enhances the model's ability to capture complex relationships in the data through learnable spline-based functions. As shown, KANs have a strong theoretical foundation for approximating arbitrary functions. In contrast, CNNs rely on convolution operations to extract spatial features. The convolution of an image $f$ with a filter $g$ is given by:

\begin{align}
(f * g)(x, y) = \sum_{u=-k}^{k} \sum_{v=-k}^{k} f(x-u, y-v) g(u, v)
\end{align}

where $f(x, y)$ represents the pixel intensity at location $(x, y)$ in a 2D image and $g(u, v)$ is the convolution filter (also called a kernel) used to detect specific patterns or features in the image. The filter has dimensions $(2k + 1) \times (2k + 1)$, meaning it covers a small region around the central pixel in the image (e.g., a 3x3 or 5x5 filter). CNNs excel at detecting local spatial patterns (e.g., edges) but assume translational invariance. There is strong evidence suggesting that CNNs can struggle to capture global dependencies in data due to their inherent local receptive fields~\cite{li2024combining, s24010274}. KANs, however, impose no such assumptions, allowing them to possibly capture both local and global patterns, leading to better generalization in some cases. In other words, KANs model both local and global dependencies simultaneously through their function decomposition. This allows KAN-based autoencoders to reconstruct data with higher precision, especially in cases where global structure is important.

On the other hand, since KAN layers use spline functions they are able to to adaptively model nonlinear relationships in data with a fine level of detail. In contrast, CNNs rely on activation functions like ReLU to introduce nonlinearity. Also, CNNs typically require explicit regularization techniques, such as batch normalization or dropout, to prevent overfitting. Therefore, KAN autoencoders may potentially provide a more natural and flexible way of modeling complex relationships.

\section{Results}

In this section, we present the results of our experiments on three datasets: MNIST, CIFAR-10, and SVHN. We compare the performance of autoencoders built with CNNs and KANs.

\subsection{Datasets}

\textbf{MNIST:} The MNIST dataset consists of 70,000 grayscale images of handwritten digits, with each image having a resolution of 28x28 pixels. The dataset is divided into 60,000 training images and 10,000 test images. Each image belongs to one of ten classes, representing the digits 0 through 9. The simplicity and standardized format of MNIST make it a popular benchmark for evaluating image processing algorithms.

\textbf{SVHN:} The Street View House Numbers (SVHN) dataset contains 600,000 color images of house numbers extracted from Google Street View images. Each image is 32x32 pixels and includes multiple digits, but the task typically involves classifying the digit at the center of the image. The dataset is split into 73,257 training images, 26,032 test images, and 531,131 additional training images. SVHN is more complex than MNIST due to its real-world origins and multi-digit context.

\textbf{CIFAR-10:} The CIFAR-10 dataset comprises 60,000 color images of size 32x32 pixels, divided into 10 classes: airplane, automobile, bird, cat, deer, dog, frog, horse, ship, and truck. Each class contains 6,000 images, with 50,000 images used for training and 10,000 for testing. The dataset poses a more challenging problem compared to MNIST due to its higher variability and color information.

\subsection{Settings and Strudcture}

Both types of autoencoders were evaluated on the MNIST, CIFAR-10, and SVHN datasets. The autoencoders were trained for 10 epochs using the AdamW optimizer with a learning rate of 1e-3 and a weight decay of 1e-4. We used the MSE loss function to measure reconstruction accuracy. The performance of the autoencoders was assessed based on the reconstruction loss on the test sets. Our AE-KAN also includes two extra dense layers compared to AE-CNN (see Fig.~\ref{fig:structure}). Additionally, we trained a KNN classifier on the latent representations learned by the autoencoders to evaluate the quality of these representations for downstream classification tasks. The accuracy and F1-score of the KNN classifier were used as additional metrics.

The KAN model used in our codes uses the efficient KAN implemantion~\cite{blealtan2023efficientkan}. The performance issue in the original implementation arises from the need to expand all intermediate variables to perform activation functions, requiring the input tensor to have the shape $(\text{batch\_size}, \text{out\_features}, \text{in\_features})$. Since all activation functions are linear combinations of a fixed set of basis functions, specifically B-splines, the computation can be reformulated by activating the input with the basis functions and then linearly combining them. This reformulation reduces memory cost and simplifies the computation to a matrix multiplication, and it works naturally for both the forward and backward pass. The original KAN proposed an L1 regularization defined on input samples, which requires non-linear operations on the $(\text{batch\_size}, \text{out\_features}, \text{in\_features})$ tensor and is incompatible with the reformulation. Efficient KAN replaces this with an L1 regularization on the weights, which is common in neural networks and compatible with the new formulation. 

In a KAN, the total number of trainable parameters is determined by the number of 
activation functions in close relation to the architecture of the network defined by 
the numbers of the input nodes ($N_i$), of the hidden nodes in each hidden layer 
($N_{h1}$, $N_{h2}$, \ldots, $N_{hj}$), and of the output nodes ($N_o$). The 
structural complexity of the KAN is then determined by the number of activation 
functions ($N_a$), expressed as:

\begin{align}
    N_a = (N_i \times N_{h1}) + (N_{h1} \times N_{h2}) + \cdots + (N_{hj} \times N_o).
\end{align}

Consider the numeric training phase of KAN. Each activation function within the KAN 
is parameterized by a B-spline curve represented as a linear combination of the basis 
functions. Each B-spline curve is characterized 
by $(G + K)$ trainable coefficients. The total number of trainable parameters in a KAN 
is then given by:

\begin{align}
    (G + K) \times N_a,
\end{align}

which gives a direct relationship between the network's architecture and its trainable 
parameters. Increasing the number of hidden layers or nodes can significantly impact 
the total number of parameters, influencing the network's capacity and complexity of 
the learned representations. 

The architecture of the KAN autoencoder consists of an encoder and a decoder. The encoder is composed of a KAN layer, followed by a ReLU activation and a dense layer. This structure transforms the input size ($\text{input\_size}$) to a hidden size ($\text{hidden\_size}$), and subsequently to the bottleneck size ($\text{bottleneck\_size}$). Specifically, the encoder can be represented as:

\begin{multline} \label{eq:enc}
\text{Encoder} = \text{KAN} + \text{ReLU} + \text{Dense} = \\
(\text{input\_size} \times \text{hidden\_size}) + \text{ReLU} + (\text{hidden\_size} \times \text{bottleneck}),    
\end{multline}

e.g., $(784 \to 8) + \text{ReLU} + (8 \to 18)$. The decoder mirrors this process, starting with a dense layer, followed by a ReLU activation, and ending with a KAN layer. The decoder transforms the bottleneck size back to the hidden size and finally to the original input size, as follows:

\begin{multline}  \label{eq:dec}
\text{Decoder} = \text{Dense} + \text{ReLU} + \text{KAN} = \\
(\text{bottleneck} \times \text{hidden\_size}) + \text{ReLU} + (\text{hidden\_size} \times \text{input\_size}),
\end{multline}

e.g., $(18 \to 8) + \text{ReLU} + (8 \to 784)$. 

\begin{figure}
    \centering
    \includegraphics[width=1\linewidth]{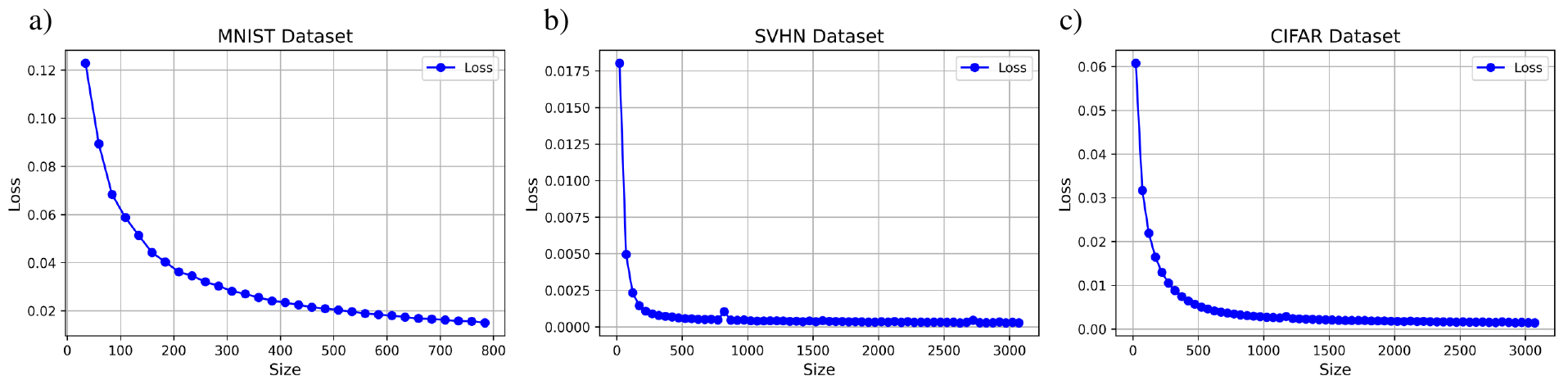}
    \caption{MSE reconstruction loss versus bottleneck size for the KAN autoencoder across three datasets: a) MNIST, b) SVHN, and c) CIFAR-10. Let $\text{hidden\_size} = \text{bottleneck\_size}$ in Eqs.~\ref{eq:enc} and \ref{eq:dec}. For the MNIST dataset, a bottleneck size of 150 achieves excellent performance, while 50 provides good performance. In the case of CIFAR-10, a bottleneck size of 500 yields excellent performance, with 200 being very good. For SVHN, a bottleneck size of 500 is excellent, and 200 shows good reconstruction.}
    \label{fig:elbow}
\end{figure}

\begin{figure}
    \centering
    \includegraphics[width=1\linewidth]{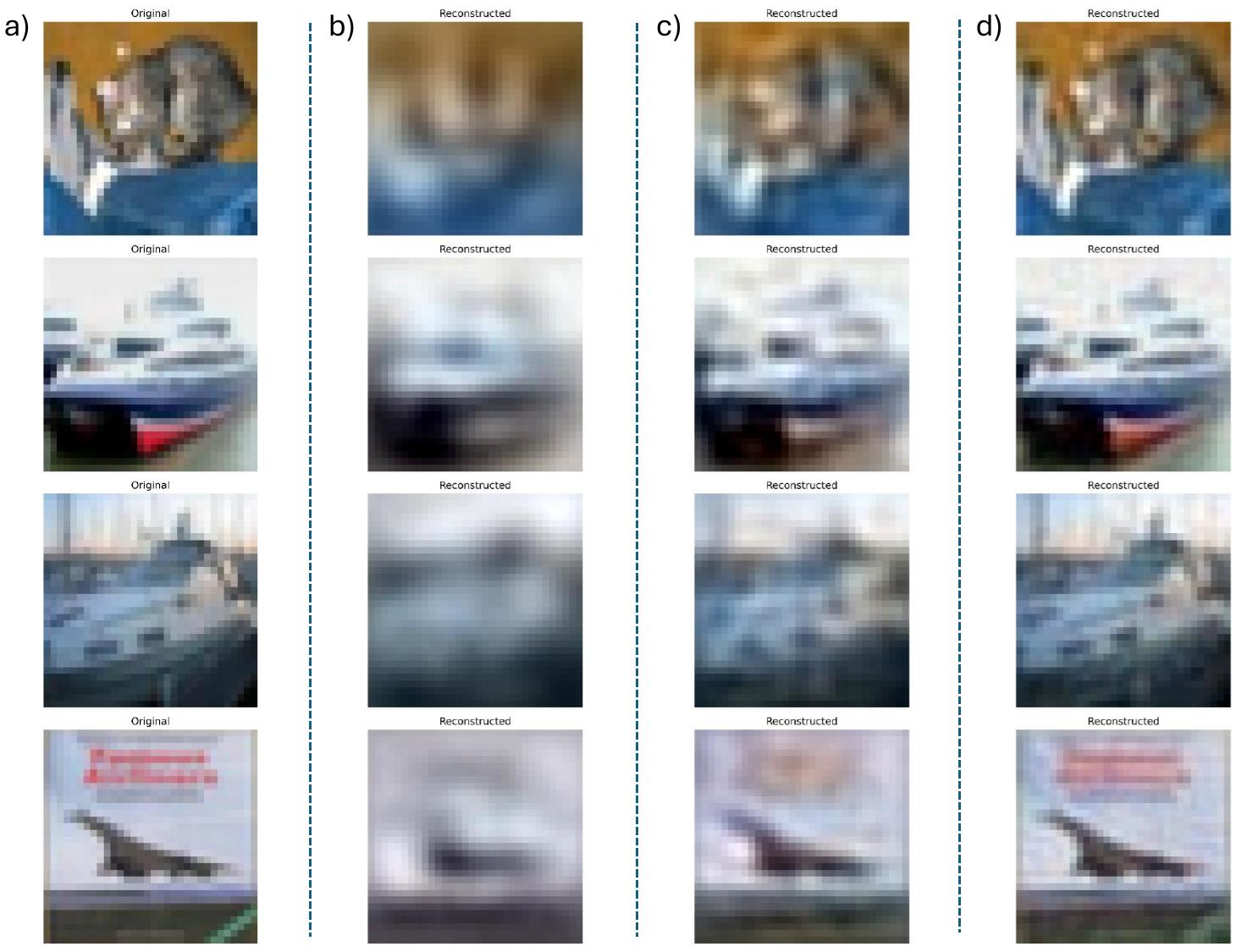}
    \caption{Reconstruction of Compressed Images from the CIFAR Dataset with Different Bottleneck Sizes. Let $\text{hidden\_size} = \text{bottleneck\_size}$ in Eqs.~\ref{eq:enc} and \ref{eq:dec}. a) Original samples b) Reconstructed samples with $\text{bottleneck\_size} = 50$, c) Reconstructed samples with $\text{bottleneck\_size} = 150$, and d) Reconstructed samples with $\text{bottleneck\_size} = 500$.}
    \label{fig:bott_comp}
\end{figure}

As illustrated in Fig.~\ref{fig:elbow}, we evaluated the performance of the KAN autoencoder considering the MSE reconstruction loss analyzed as a function of the bottleneck size ($\text{hidden\_size} = \text{bottleneck\_size}$ in Eqs.~\ref{eq:enc} and \ref{eq:dec}). For the MNIST dataset, we observed that a bottleneck size of 150 produced the best reconstruction results, achieving an excellent MSE score. A smaller bottleneck size of 50 also provided reasonable performance, though the reconstruction quality was slightly lower compared to the larger bottleneck. In the case of CIFAR-10, we found that a bottleneck size of 500 yielded the best performance, resulting in an excellent MSE. Interestingly, a smaller bottleneck size of 200 also performed very well, though slightly less optimal compared to 500. For the SVHN dataset, the autoencoder showed similar behavior to CIFAR-10, where a bottleneck size of 500 achieved excellent performance, while a smaller size of 200 offered good performance but with a higher MSE compared to the larger bottleneck. Figure~\ref{fig:bott_comp} compares the reconstruction of compressed images from the CIFAR-10 dataset with varying bottleneck sizes. For these reconstructions, we set the hidden size equal to the bottleneck size, as defined in Eqs.~\ref{eq:enc} and \ref{eq:dec}. The figure illustrates: a) original samples, b) reconstructions with a bottleneck size of 50, c) reconstructions with a bottleneck size of 150, and d) reconstructions with a bottleneck size of 500. As observed, larger bottleneck sizes lead to clearer and more defined reconstructions. This is because a larger bottleneck allows the model to retain more information from the original images, resulting in higher fidelity reconstructions. In contrast, smaller bottleneck sizes tend to produce more blurred images. The limited capacity of the smaller bottleneck compresses the information too much, leading to a loss of essential details and features present in the original samples.

Consider a KAN autoencoder with a grid size of 5 with the autoencoder bottleneck size of 4, using the MNIST dataset as input (dimension $28 \times 28$). The architecture includes two dense layers each with an dense size of 8 (see Fig.~\ref{fig:structure}). The total number of parameters in this configuration is 62,796, which includes the parameters of both the encoder and decoder KANs, as well as those of the dense layers in the encoder and decoder. The total number of trainable parameters in a KAN autoencoder can be calculated using the following formula:

\begin{multline}
N_{\text{params}} = N_{\text{enc\_params}}+ N_{\text{enc\_dense\_params}}+N_{\text{dec\_params}}+ N_{\text{dec\_dense\_params}} \\
=\text{grid} \times \text{input\_size} \times \text{hidden\_size} + \text{hidden\_size} \times (\text{bottleneck\_size} + 1)\\
+ \text{grid} \times \text{hidden\_size} \times \text{input\_size} + \text{bottleneck\_size} \times (\text{hidden\_size} + 1)      
\end{multline}

\subsection{Results and Discussion}

The results of our experiments are summarized in Table~\ref{table:MNIST_results}.

\begin{table}[h]
    \centering
    \caption{KAN Autoencoders vs CNN Autoencoders - Reconstruction and Classification Results.}
    \label{table:MNIST_results}
    \begin{tabular}{lcccccccc}
        \toprule
        Model & Dataset & Bottle. & Param. & Time & Param/Time & Loss & Acc. & F1 \\
        \midrule\midrule
        AE-CNN & MNIST & $2\times2$ & 617 & \textbf{39.24} & 15.72 & 0.2667 & 0.74 & 0.74 \\
        AE-KAN & MNIST & $4$ & 62796 & 40.82 & \textbf{1538.36}  & \textbf{0.2170} & \textbf{0.76} & \textbf{0.75} \\
        \midrule
        AE-CNN & MNIST & $3\times3$ & 818 & 44.21 & 18.50  & 0.2288 & 0.82 & 0.82 \\
        AE-KAN & MNIST & $9$ & 62881 & \textbf{43.14} & \textbf{1457.60}  & \textbf{0.2024} & \textbf{0.84} & \textbf{0.84} \\
        \midrule
        AE-CNN & MNIST & $6\times3$ & 891 & \textbf{43.47} & 20.49  & \textbf{0.1722} & \textbf{0.90} & \textbf{0.90} \\
        AE-KAN & MNIST & $18$ & 63034 & 42.94 & \textbf{1467.95}  & 0.2238 & 0.78 & 0.77 \\
        \midrule
        AE-CNN & SVHN & $8\times4$ & 3509 & \textbf{71.70} & 48.94  & 0.0251 & 0.67 & 0.66 \\
        AE-KAN & SVHN & $32$ & 9.97e5 & 72.70 & \textbf{13713.89}  & \textbf{0.0129} & \textbf{0.67} & \textbf{0.68} \\
        \midrule
        AE-CNN & SVHN & $12\times4$ & 6854 & \textbf{71.53} & 95.81  & 0.0167 & 0.67 & 0.66 \\
        AE-KAN & SVHN & $48$ & 1.00e6 & 73.07 & \textbf{13685.50}  & \textbf{0.0089} & \textbf{0.69} & \textbf{0.68} \\
        \midrule
        AE-CNN & SVHN & $16\times4$ & 30731 & \textbf{71.88} & 427.53  & 0.0114 & 0.67 & 0.66 \\
        AE-KAN & SVHN & $64$ & 1.01e6 & 73.30 & \textbf{13778.99}  & \textbf{0.0082} & \textbf{0.69} & \textbf{0.68} \\
        \midrule
        AE-CNN & CIFAR & $8\times4$ & 6709 & \textbf{43.32} & 154.87  & 0.0659 & 0.55 & 0.56 \\
        AE-KAN & CIFAR & $32$ & 1.99e6 & 44.09 & \textbf{45227.27}  & \textbf{0.0497} & \textbf{0.57} & \textbf{0.58} \\
        \midrule
        AE-CNN & CIFAR & $12\times4$ & 6854 & \textbf{44.68} & 153.40  & 0.0523 & 0.57 & 0.57 \\
        AE-KAN & CIFAR & $48$ & 2.00e6 & 45.99 & \textbf{43487.71}  & \textbf{0.0402} & \textbf{0.58} & \textbf{0.58} \\
        \midrule
        AE-CNN & CIFAR & $16\times4$ & 6999 & \textbf{45.82} & 152.74  & 0.0414 & 0.57 & 0.57 \\
        AE-KAN & CIFAR & $64$ & 2.02e6 & 46.70 & \textbf{43254.81}  & \textbf{0.0341} & \textbf{0.59} & \textbf{0.59} \\
        \bottomrule
    \end{tabular}
\end{table}

\begin{figure}
    \centering
    \includegraphics[width=1\linewidth]{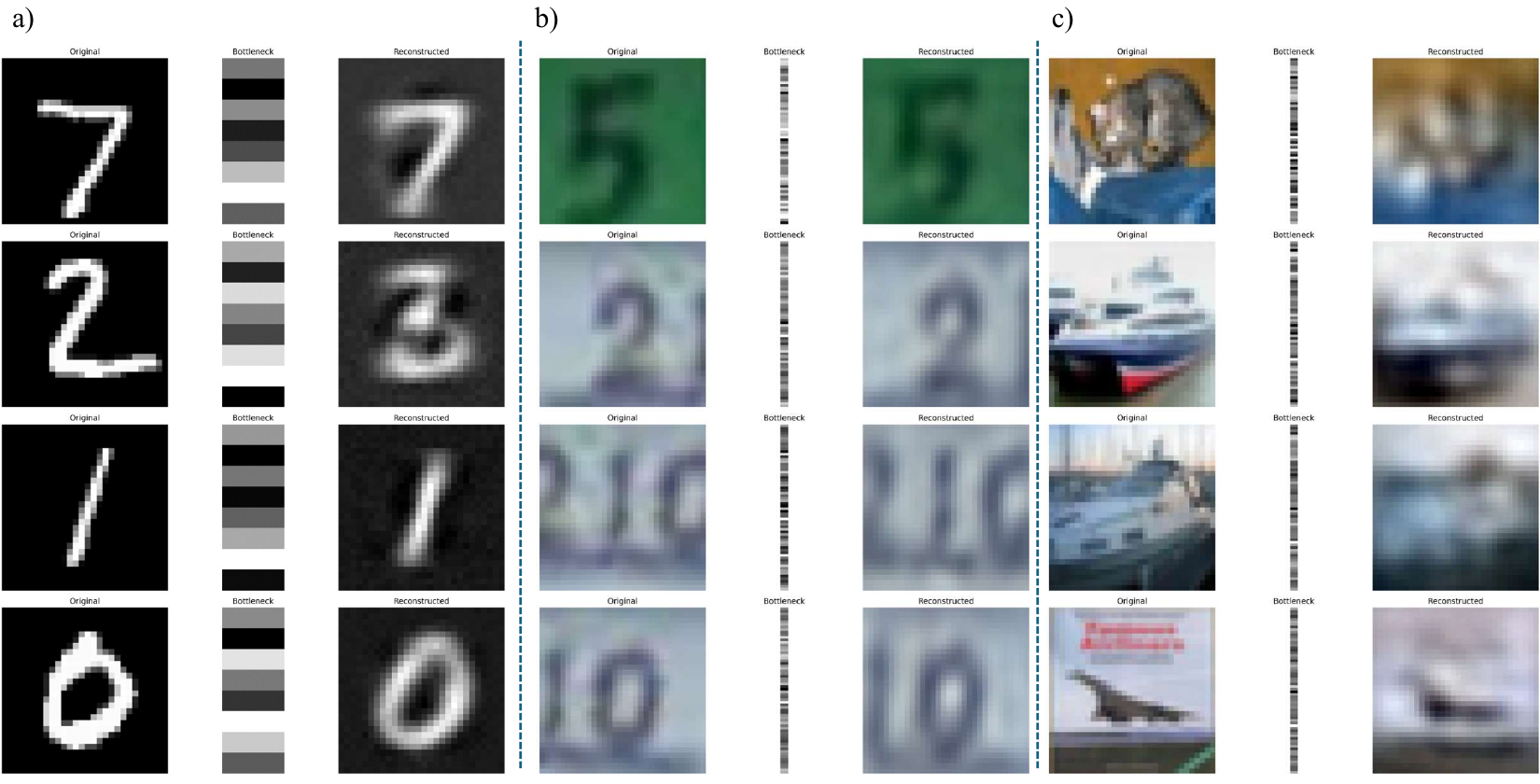}
    \caption{KAN Autoencoders Image Reconstruction Results: a) MNIST Dataset ($\text{bottleneck\_size} = 8$) b) SVHN Dataset ($\text{bottleneck\_size} = 64$) c) CIFAR-10 Dataset ($\text{bottleneck\_size} = 64$).}
    \label{fig:recon}
\end{figure}

\begin{figure}
    \centering
    \includegraphics[width=1\linewidth]{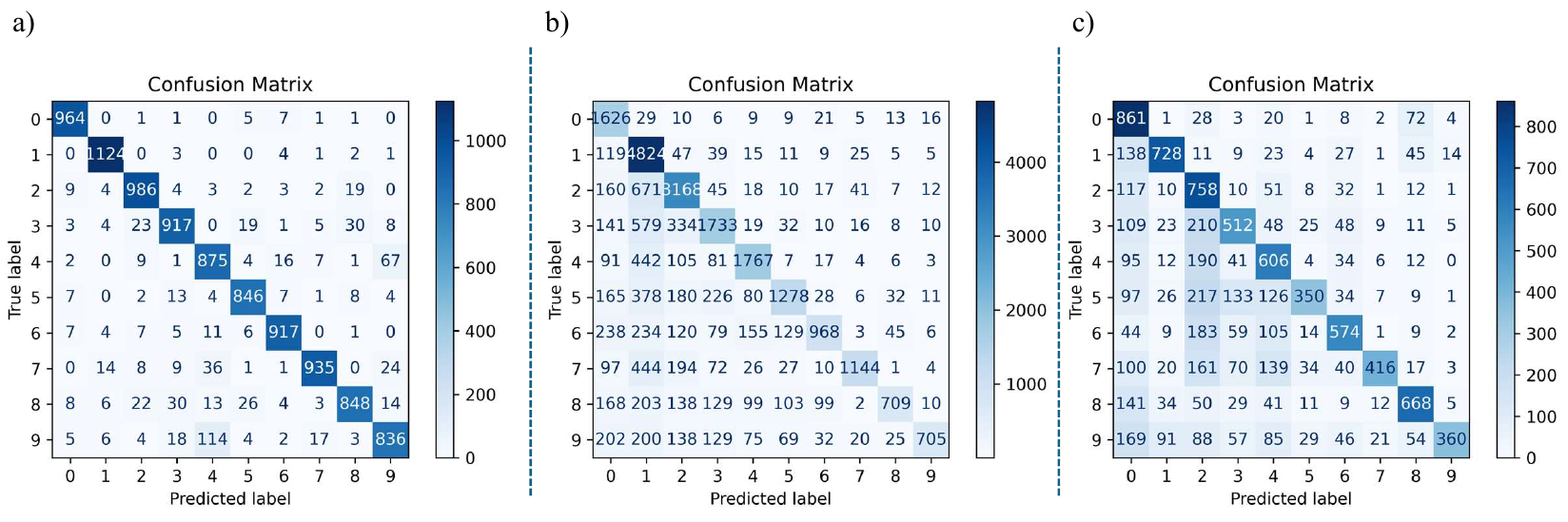}
    \caption{KNN Classification Results Given the Latent Representations of the KAN Autoencoders: a) MNIST Dataset ($\text{bottleneck\_size} = 8$) b) SVHN Dataset ($\text{bottleneck\_size} = 64$) c) CIFAR-10 Dataset ($\text{bottleneck\_size} = 64$).}
    \label{fig:class}
\end{figure}

For the MNIST dataset (see the reconstruction sample with $\text{bottleneck\_size} = 8$ in Figs.~\ref{fig:recon}a and \ref{fig:class}a), the performance of AE-KAN and AE-CNN varies across different bottleneck configurations. At the smallest bottleneck configuration (2x2 for AE-CNN and 4 for AE-KAN), AE-KAN demonstrates superior reconstruction loss (0.2170) compared to AE-CNN (0.2667). This suggests that AE-KAN is more effective at reconstructing the original data from a compressed representation, potentially due to the unique structure of KANs where activation functions on edges might provide richer transformations. Additionally, AE-KAN shows slightly higher accuracy and F1-score, indicating better classification performance. However, this comes at the cost of significantly more parameters (62,796 for AE-KAN versus 617 for AE-CNN), suggesting that AE-KAN's improved performance may partly stem from its higher complexity and capacity. As the bottleneck size increases to 3x3 for AE-CNN and 9 for AE-KAN, AE-KAN continues to outperform AE-CNN in terms of reconstruction loss (0.2024 vs. 0.2288) and achieves marginally better accuracy and F1-scores. Interestingly, AE-KAN completes this task slightly faster (43.14 seconds) than AE-CNN (44.21 seconds), despite having a higher parameter count (62,881 vs. 818). However, with the largest bottleneck configuration (6x3 for AE-CNN and 18 for AE-KAN), AE-CNN outperforms AE-KAN. AE-CNN achieves a reconstruction loss of 0.1722 and accuracy and F1-scores of 0.90, compared to AE-KAN's 0.2238 reconstruction loss and 0.78 accuracy and F1-scores. This reversal indicates that AE-CNN can leverage its structure more effectively for larger bottlenecks.

On the SVHN dataset (see the reconstruction sample with $\text{bottleneck\_size} = 64$ in Figs.~\ref{fig:recon}b and ~\ref{fig:class}b) as well as the CIFAR-10 dataset (see the reconstruction sample with $\text{bottleneck\_size} = 64$ in Figs.~\ref{fig:recon}c and ~\ref{fig:class}c), the pattern remains consistent with AE-KAN generally outperforming AE-CNN in terms of reconstruction loss across all bottleneck sizes. For instance, with a bottleneck size of 8x4 for AE-CNN and 32 for AE-KAN, AE-KAN achieves a reconstruction loss of 0.0129 compared to AE-CNN's 0.0251.

The results show that AE-KAN consistently outperforms AE-CNN in both reconstruction accuracy and classification, especially on datasets with complex features such as SVHN and CIFAR-10. The key difference lies in the structural adaptation of KANs, which offers a greater representation capacity, leading to improved performance metrics. However, this benefit comes at the cost of increased model complexity and a higher parameter count, potentially raising computational demands. Moreover, we need to emphasize that AE-KAN includes two extra dense layers, an addition that AE-CNN does not require since its bottleneck size can be tuned using multiple CNN layers. This difference may also partially contribute to AE-KAN's superior performance.

\section{Discussion}

In this study, we investigated the application of KANs in the context of autoencoder architectures for image representation tasks. Our experiments focused on comparing KAN-based autoencoders with traditional CNN-based autoencoders across three benchmark datasets: MNIST, CIFAR-10, and SVHN. The primary objectives were to evaluate reconstruction accuracy and assess the quality of learned representations using classification tasks. To ensure a fair comparison between the CNN and KAN autoencoders, we maintained identical training conditions for both models. This included using the same optimizer, learning rate, weight decay, number of epochs, and batch size. Our findings indicate that KAN-based autoencoders performed competitively compared with CNN-based autoencoders in terms of reconstruction accuracy on all three datasets. Specifically, in some cases KANs achieved lower MSE losses, suggesting their ability to capture more informative and discriminative features from the input data. Furthermore, the latent representations learned by KAN-based autoencoders improved classification accuracy and F1-scores when evaluated with a KNN classifier.

While the employed datasets provide a good range of complexity, they are still limited to image classification tasks. The generalizability of KAN-based autoencoders to other domains, such as text or time-series data, remains unexplored. Although we observed improved performance in our experiments, the training time and resource requirements for KAN-based models were higher compared to traditional CNN-based autoencoders. For instance, KAN-based reinforcement learning models have been shown to perform comparably to MLP-based models but with fewer parameters. This has sparked debates on whether KAN's added complexity is justified, given that similar performance levels are achievable with simpler architectures like MLPs. Researchers are still evaluating whether KAN offers enough distinct advantages in various domains, such as physics and offline reinforcement learning~\cite{guo2024kanvsmlpoffline}. This could be a limiting factor for large-scale or real-time applications. Also, while we discuss the potential for improved interpretability with KANs, our study did not include formal metrics or methods to quantify this aspect. Future work should focus on developing and applying interpretability metrics to rigorously evaluate the interpretability of learned representations.

Future work should explore applying KAN-based autoencoders to real-world problems, particularly in fields requiring high interpretability and accuracy, such as medical imaging, finance, and autonomous systems. Case studies and domain-specific evaluations would provide a more comprehensive understanding of their impact. Also, investigating hybrid architectures that combine the strengths of KANs with other neural network architectures, such as Recurrent Neural Networks (RNNs) or Graph Neural Networks (GNNs), could lead to novel models with enhanced capabilities.

\begin{ack}

This work was supported by AFOSR under Grant No.~FA9550-21-1-0438.

\end{ack}

\subsection*{Code Availability}

The code used for experiments in this study is publicly available at GitHub~\cite{moradi2023kanautoencoders}.

\bibliography{ref.bib}
\end{document}